\documentclass[letterpaper, 10 pt, conference]{ieeeconf}  

\IEEEoverridecommandlockouts                              

\usepackage{xcolor}

\usepackage{graphicx} 
\usepackage{amsmath} 
\usepackage{booktabs}
\usepackage{multirow}

\title{\LARGE \bf
Bridging the Awareness Gap: Socially Mediated State Externalization for Transparent Distributed Home Robots
}

\author{
Wenzheng Zhao$^{1}$, Manideep Duggi$^{1}$, Fengpei Yuan$^{1}$%
\thanks{*This work has been submitted to IROS 2026 for review.}%
\thanks{$^{1}$All authors are with Faculty of Robotics Engineering Department, Worcester Polytechnic Institute, 100 Institute Rd, Worcester, MA, USA
{\tt\small [wzhao8@wpi.edu, mduggi@wpi.edu, fyuan3@wpi.edu](mailto:fyuan3@wpi.edu)}}%
}

\begin{document}

\maketitle
\thispagestyle{empty}
\pagestyle{empty}

\begin{abstract}

Distributed multi-robot systems for the home often require robots to operate out of the user's sight, creating a state awareness gap that can diminish trust and perceived transparency and control. This paper investigates whether real-time, socially mediated state externalization can bridge this gap without compromising task performance. We developed a system where a co-located social mediator robot (Pepper) externalizes the hidden execution states of an out-of-sight mobile manipulator (Stretch~3) for voice-driven object retrieval and delivery, where task-level states are synchronized and externalized through verbal updates and visual progress display. 
In a counterbalanced within-subject study ($N=30$), we compared a baseline of \textit{Autonomous Hidden Execution} against \textit{Socially Mediated State Externalization}. Our results show that externalization significantly increases user task-focused attention (from 15.8\% to 84.6\%, $p<.001$) and substantially improves perceived perspicuity, dependability, stimulation, and attractiveness (all $p<.001$). Furthermore, 83\% of participants preferred the externalized condition, and this improvement in user experience was achieved without a statistically significant increase in end-to-end task completion time ($p=.271$). The results suggest that socially mediated state externalization is an effective architectural mechanism for designing more transparent and trustworthy distributed robot systems, ultimately enhancing user experience without sacrificing performance in distributed home robot deployments.

\end{abstract}

\section{INTRODUCTION}
When a robot performs a task out of the user's line of sight—retrieving an object from another room while the user waits—the user cannot observe progress, intermediate states, or failures. We term this the \textit{state awareness gap}: spatial separation removes the implicit cues that normally support predictability and transparency in human-robot interaction. As domestic robots evolve from single, visible units toward distributed multi-robot ecosystems, this gap will become increasingly common. A mobile manipulator may execute tasks in one space while a separate robot remains co-located with the user, enabling modular deployments but introducing a fundamental interaction challenge.

Designing for this gap presents a fundamental tension. Fully autonomous systems often prioritize seamless operation, automatically handling failures and recovery without interrupting the user. Yet, externalizing internal states—such as task progression and, when they occur, failure—may improve perceived control, understanding, and trust. Whether distributed home robot systems should hide internal execution and recovery processes or explicitly externalize them remains an open design question. 

\begin{figure}[htp]
  \centering
  \includegraphics[width=\linewidth]{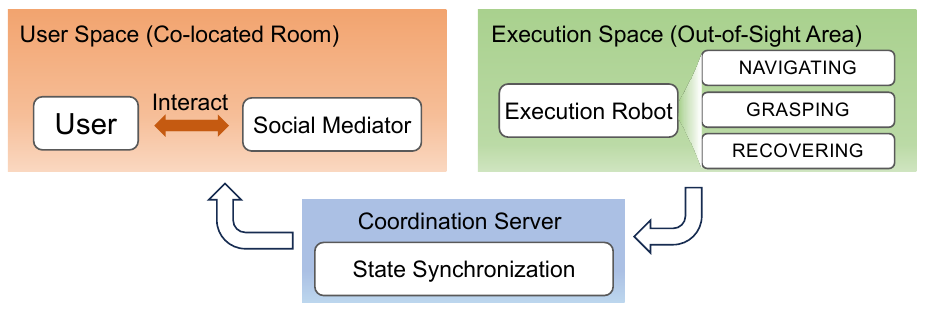}
  \caption{Overview of the distributed execution architecture and state externalization framework. The user interacts with a co-located social mediator, while task execution is performed by a mobile manipulation robot operating in a physically separated workspace. A coordination server synchronizes task-level execution states (e.g., NAVIGATING, GRASPING, RECOVERING) and externalizes them to the user through verbal and visual feedback. This architecture enables real-time state awareness despite spatial separation.}
  \label{fig:system}
\end{figure}

In this paper, we investigate whether real-time, socially mediated state externalization can bridge the awareness gap without compromising task performance. We implement a two-agent architecture (Fig.~\ref{fig:system}): an execution robot performs object retrieval and delivery tasks in a separate space, while a co-located humanoid social robot (Pepper) acting as a social mediator externalizes the execution state through verbal updates and a visual progress display. In a controlled within-subject study (\textit{N} = 30), we compare this approach against a baseline of autonomous hidden execution where task states are not externalized.

\textbf{Our contributions are as follows:}
\begin{itemize}
    \item Empirical evidence that socially mediated state externalization significantly improves perceived transparency, dependability, and engagement (all $p<.001$) while increasing task-focused attention from 15.8\% to 84.6\%.
    \item A distributed architecture and semantic state abstraction mechanism that enables real-time state externalization across physically separated agents.
    \item Design implications for transparency in distributed domestic robot systems, drawn from empirical results of objective performance and subjective measures.

\end{itemize}

\section{RELATED WORK}

\textbf{Transparency and Explainability in Human–Robot Interaction.}
Transparency and explainability have become increasingly important in autonomous robotic systems, particularly in human-centered environments. Prior work has shown that making robot intentions, internal states, or task progress visible can improve user trust, predictability, and perceived reliability~\cite{dragan2013legibility}. Approaches to transparency range from visual motion cues and embodied progress displays to verbal explanations of robot reasoning. For example,~\cite{schott2023literature} provide a structured overview of how transparency is communicated in HRI research and highlight verbal and visual explanations as common strategies to improve understanding.
More recent work has proposed empirical measures of perceived transparency, suggesting that explainability, legibility, and predictability are core dimensions that contribute to how humans interpret robot behavior~\cite{angelopoulos2025measuring}. However, most prior studies assume co-located interaction, where users can directly observe robot motion. Less attention has been given to transparency challenges arising from spatially distributed robotic systems in which task execution is not directly visible.

\textbf{Autonomous Recovery and User Awareness.}
Autonomous robots commonly incorporate failure detection and recovery mechanisms to ensure robust task execution in real-world environments~\cite{kaelbling2011hierarchical}. In many deployed systems, recovery processes are handled internally to minimize user interruption and preserve interaction fluency. While such hidden recovery strategies can improve efficiency and smoothness of interaction, prior research suggests that reduced visibility of system uncertainty may negatively affect user situation awareness and perceived control~\cite{chen2018situation}. Research in shared autonomy and human-in-the-loop systems further indicates that involving users in decision-making during uncertain or failure states can improve perceived transparency and trust~\cite{nikolaidis2017human,gopinath2016human}. However, most existing work focuses on collaborative or co-located interaction scenarios. The tradeoff between seamless autonomous recovery and explicit failure externalization remains underexplored in distributed domestic robot settings, where users cannot directly observe task execution.

\textbf{Social Mediation in Human-Robot Interaction.}
Social robots have been investigated as mediators that facilitate communication between autonomous systems and users~\cite{breazeal2003toward}. Prior work demonstrates that co-located social agents can provide explanations, task updates, and engagement cues that enhance user experience~\cite{sidner2005explorations,mutlu2009footing}. Social mediation has been explored in collaborative robots, service robots, and assistive contexts to improve user engagement and clarity of interaction~\cite{mataric2007personalized, fasola2013socially}. Additionally, research on explainable robot behavior suggests that verbal and mediator-delivered visual explanations can improve perceived transparency and predictability~\cite{hayes2017improving, halilovic2025explainable}. However, existing studies primarily focus on improving interaction quality rather than examining the role of social mediation in bridging state awareness gaps created by spatially separated task execution.

In contrast to prior work, this paper investigates how real-time state externalization through a social mediator affects user perception in distributed home robot systems where execution occurs outside direct user observation.

\section{SYSTEM DESIGN AND STATE EXTERNALIZATION FRAMEWORK}

This section describes the distributed robotic architecture and the proposed execution state externalization mechanism. Our goal is to formalize how execution states in spatially separated robotic systems can be modeled, synchronized, and externalized to users in real time.

\subsection{Distributed Execution Architecture}

\begin{figure*}[htp]
  \centering
  \includegraphics[width=0.99\linewidth]{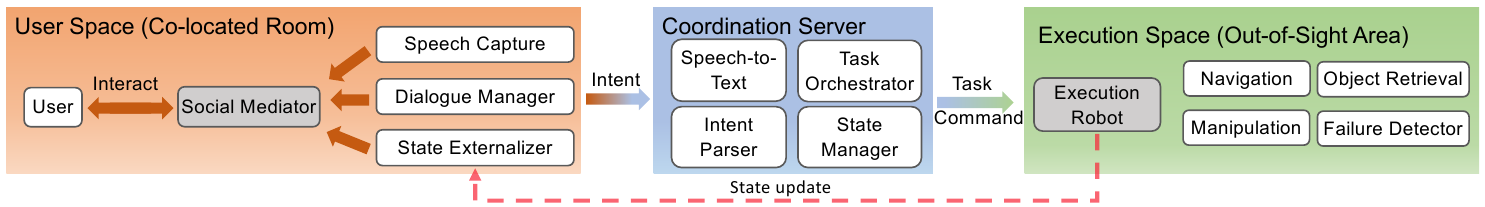}
  \caption{
System architecture of the distributed robot interaction framework. 
User voice commands are captured by the Pepper social mediator, which 
forwards structured intents to a coordination server. The server 
dispatches task commands to the Stretch mobile manipulator for 
navigation and object retrieval. During execution, the robot publishes 
task-level state updates (e.g., NAVIGATING, SEARCHING, GRASPING), which 
are externalized to the user through Pepper via verbal and visual 
feedback. When a failure is detected, a closed-loop recovery process 
is triggered: the failure event is externalized to the user, user 
confirmation is collected, and a recovery command is issued to the 
execution robot.
}
  \label{fig:Architecture}
\end{figure*}

As shown in Fig.~\ref{fig:Architecture}, we implement a distributed two-agent robotic system consisting of:
\textbf{(1) an execution robot}, a mobile manipulation platform responsible for object retrieval and delivery tasks that operates in a separate physical space from the user, and
\textbf{(2) a social mediator}, a co-located social robot that remains in the same room as the user and serves as the embodied mediator for interaction and state communication.

The execution robot performs navigation, object search, grasping, and delivery behaviors autonomously. Importantly, these actions occur outside the user’s direct line of sight. As a result, the user cannot infer task progress through visual observation alone.

To bridge this separation, the system employs a real-time communication layer in which the execution robot publishes task-level state updates. The social mediator subscribes to these updates and externalizes them through human-readable communication channels (visual and auditory feedback). This design enables the decoupling of task execution from user interaction while maintaining state awareness.

\subsection{Execution State Modeling}

To enable consistent state externalization, we formalize a task-level execution state model. Rather than exposing low-level controller signals, we define a discrete set of semantically meaningful execution states summarized in Table~\ref{tab:execution_states}.

\begin{table}[t]
\centering
\caption{Task-level execution state definitions.}
\label{tab:execution_states}
\begin{tabular}{p{1.6cm} p{6.2cm}}
\toprule
\textbf{State} & \textbf{Description} \\
\midrule
IDLE & The robot is awaiting task assignment. \\

NAVIGATING & The robot is moving toward a target location. \\

SEARCHING & The robot is locating the requested object. \\

GRASPING & The robot is attempting object manipulation. \\

FAILED & A navigation or manipulation failure has been detected. \\

RECOVERING & The robot is executing a recovery strategy. \\

DELIVERING & The robot is returning to the user with the object. \\
\bottomrule
\end{tabular}
\end{table}

This abstraction layer serves two purposes. First, it standardizes state representation across the distributed system. Second, it provides a human-interpretable medium for communication. By mapping low-level controller events to task-level semantic states, we ensure that the externalized information remains meaningful to users without exposing unnecessary system complexity.

\subsection{State Externalization Mechanism}

The core contribution of this work lies in how execution states are externalized to the user. In the baseline condition (Autonomous Hidden Execution), state transitions and recovery behaviors occur internally within the execution robot. Failures are handled autonomously, and recovery attempts are not communicated to the user unless the task ultimately fails.

In the proposed condition (Socially Mediated State Externalization), execution state transitions are communicated in real time to the social mediator. The mediator translates state updates into human-readable verbal messages and visual progress cues. In the event of a failure state, the mediator explicitly informs the user of the detected issue and requests input on whether to reattempt the task. Upon user confirmation, the execution robot initiates a recovery strategy and transitions into the \textit{RECOVERING} state. 

This mechanism introduces a human-mediated recovery loop while preserving the underlying autonomous control pipeline. Importantly, the externalization layer does not modify low-level task execution logic; rather, it exposes task-level state transitions and enables user awareness of system uncertainty.

By isolating state externalization as the primary design variable, the framework allows us to evaluate the perceptual impact of transparency independently of task performance differences.

\subsection{System Execution and Implementation}

To support stable and interpretable state externalization, the system adopts a hybrid execution structure that separates task-level semantics from low-level control processes. At the coordination level, a central server dispatches task requests to the execution robot and monitors task progress asynchronously. This enables non-blocking communication between distributed components and ensures that task-level state updates can be propagated independently of control execution. On the execution robot, tasks progress through well-defined semantic phases (e.g., \textit{NAVIGATING} $\rightarrow$ \textit{GRASPING} $\rightarrow$ \textit{DELIVERING}). These phases correspond directly to the state model described earlier and form the basis for externalized user feedback. Internally, navigation and manipulation modules operate concurrently as independent processes. For example, the ROS~2 Nav2 stack handles localization and path planning for mobile navigation, while grasp execution is implemented using closed-loop visual servoing with an eye-in-hand depth camera. The state transitions are triggered only at task-level milestones (e.g., navigation success, grasp completion, failure detection). This separation between low-level control loops and high-level semantic states ensures that externalized information remains stable, interpretable, and synchronized across the distributed architecture.

\textbf{Hardware Platform:} The social mediator is implemented on a humanoid social robot, Pepper~\cite{pandey2018mass}, equipped with a microphone array for speech input, an onboard speaker for verbal output, and a tablet display for visual feedback. The execution platform is a Hello Robot Stretch~3~\cite{kemp2022design} mobile manipulator equipped with a differential-drive base and a telescoping arm with a dexterous wrist and gripper. For manipulation, the system employs an eye-in-hand Intel RealSense D405 depth camera mounted on the gripper (DexWrist3 configuration). For navigation, Stretch utilizes onboard sensing suitable for 2D navigation, including a planar lidar sensor, and runs ROS~2-based autonomy for localization and motion planning.


\section{EXPERIMENTAL METHODOLOGY}

This section describes the experimental setup used to evaluate the perceptual impact of execution state externalization in distributed home robot systems. The setup was designed to isolate the effect of state transparency while keeping task execution logic consistent across conditions. The system enables user-voice-driven object retrieval and delivery tasks using a distributed architecture composed of three computational roles: (i) a embodied social mediator (Pepper) for user speech interaction and audio capture, which also presents real-time task/status updates via co-located verbal and  visual progress cues on its tablet, (ii) a central coordination server responsible for speech-to-text processing, intent parsing, task orchestration, and inter-robot communication, and (iii) a Stretch~3 mobile manipulator for physical task execution. 

The coordination server converts user-spoken requests into structured task intents and dispatches them to the execution robot via a lightweight REST API. During task execution, the server continuously monitors robot state transitions and synchronizes them with the social mediator, enabling real-time state externalization.

\subsection{Experimental Conditions}

\begin{figure}[t]
  \centering
  \includegraphics[width=0.99\linewidth]{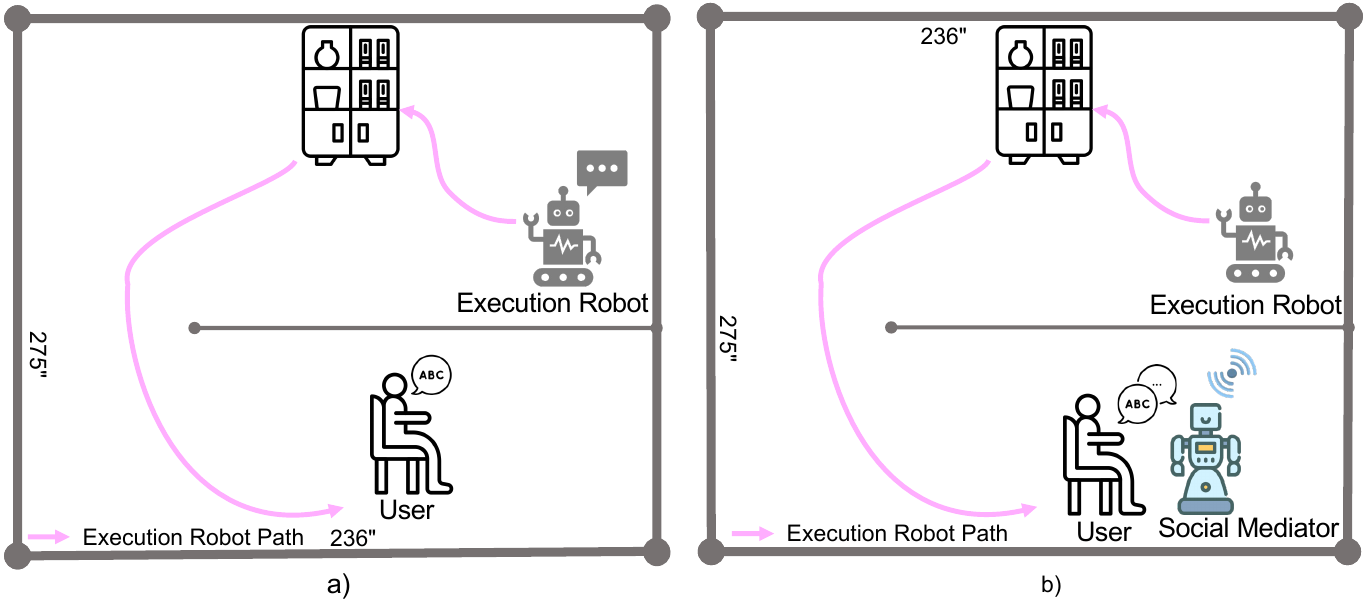}
  \caption{
Experimental layout of the distributed task execution setup under two conditions. 
(a) \textbf{Condition A}: Autonomous Hidden Execution: the execution robot operates in a physically separated workspace without communicating intermediate task states to the user. 
(b) \textbf{Condition B}: Socially Mediated State Externalization: a co-located social mediator provides real-time verbal and visual updates about the execution robot’s task states. 
In both conditions, the execution robot performs object retrieval in a separate room, while the user remains spatially separated from task execution. 
The purple arrow indicates the execution robot’s navigation path during retrieval and delivery tasks.
}
  \label{fig:expsetup}
\end{figure}

We evaluate two execution paradigms (Fig.~\ref{fig:expsetup}) that differ only in how internal task states are communicated to the user. We used a within-subject counterbalanced design. Participants were randomly assigned to one of two sequences (A $\rightarrow$ B or B $\rightarrow$ A) to mitigate order and carryover effects. 

\textbf{Condition A: Autonomous Hidden Execution.}  
In this baseline condition, the execution robot performs navigation, object retrieval, and delivery autonomously. Failure detection and recovery are handled internally. Intermediate task states and recovery attempts are not communicated to the user unless the task ultimately fails. The user remains unaware of internal execution events while the robot operates in a separate space.

\textbf{Condition B: Socially Mediated State Externalization.}  
In this condition, the execution robot publishes real-time task-level state updates to the social mediator. The mediator externalizes execution states through verbal updates and visual progress cues. When a failure state is detected, the mediator explicitly informs the user and requests confirmation before initiating a recovery attempt. Upon user confirmation, the robot executes its recovery strategy and continues the task.

Importantly, the underlying navigation, manipulation, and recovery algorithms remain identical across both conditions. The only manipulated variable is whether execution states and failure events are externalized to the user.




Each participant experienced both experimental conditions. The order of conditions was counterbalanced to mitigate potential ordering effects.

\subsection{Participants}

\begin{figure*}[t]
  \centering
  \includegraphics[width=0.99\linewidth]{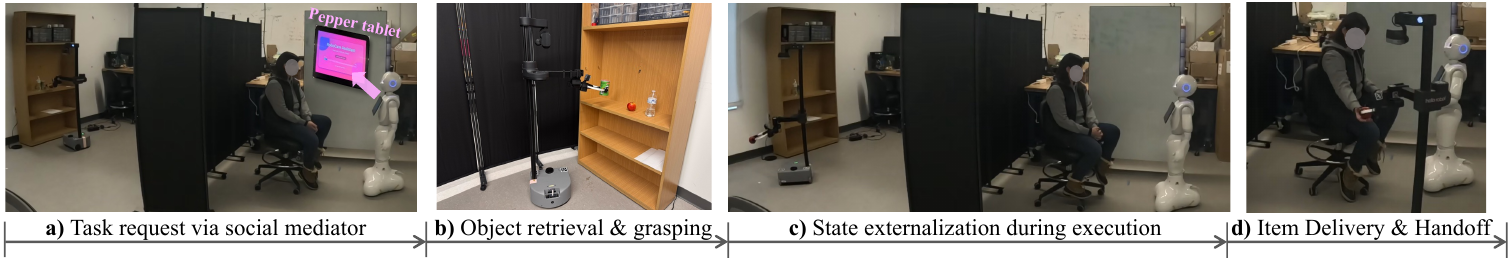}
  \caption{
Example end-to-end trial timeline for Condition B (Socially Mediated State Externalization). (a) The participant issues a request via the social mediator (Pepper) and confirms the task. (b) While the execution robot operates in a separate workspace, the mediator externalizes task progress through verbal updates and a visual progress display. (c) The execution robot retrieves the target object and returns. (d) The item is delivered to the participant, and the mediator concludes the interaction.
}
  \label{fig:realdemo}
\end{figure*}

A total of 30 participants (\textit{N}=30) were recruited from local universities, surrounding community organizations, and working professionals from diverse fields. Participants were adults with diverse academic backgrounds and varying levels of prior exposure to robotic systems, ranging from no prior experience to regular interaction with educational or service robots. Fig.~\ref{fig:realdemo} illustrates the end-to-end interaction workflow during experimental trials, including user input acquisition, real-time interaction, remote object retrieval, and final human–robot engagement. Participants were instructed to request a household object (e.g., water, snacks, or fruit) from the robot system. After the request, the execution robot navigated to a separate area, searched for the specified object, attempted grasping, and returned the object to the user.

None of the participants were informed of the experimental hypothesis prior to participation. All participants provided informed consent before the study. The experiment was conducted in accordance with institutional IRB ethical guidelines.

\subsection{Evaluation Metrics and Statistical Analysis}

To assess both system performance and user perception, we collected objective performance metrics and subjective questionnaire responses.

\textbf{Objective Performance Measures.} The {objective metrics} are summarized in Table~\ref{tab:objective_metrics}. 


\begin{table}[htp]
\centering
\caption{Objective Performance Metrics}
\label{tab:objective_metrics}
\begin{tabular}{p{3.4cm} p{4.5cm}}
\toprule
\textbf{Metric} & \textbf{Description} \\
\midrule
Task Initiation Time & Duration from system readiness for speech input to dispatch of a structured task command. \\

Execution Time & Time from task dispatch to completion (or failure), including navigation and manipulation phases. \\

End-to-End Task Time & Total duration from user request initiation to final task completion or termination. \\

Number of Grasp Attempts & Total grasp trials executed during a task session. \\

Task Success & Binary indicator of successful object retrieval and delivery. \\

Task-Focused Attention Ratio & Proportion of execution time during which participants oriented toward task-relevant areas, coded from video. \\
\bottomrule
\end{tabular}
\end{table}

\textbf{Subjective Perception Measures.} After each condition, participants completed a structured questionnaire consisting of two components.
\textit{First}, participants completed a standard 26 questions User Experience Questionnaire (UEQ)~\cite{schrepp2019design} using 7-point semantic differential scales to assess general impressions of the system, including clarity, efficiency, attractiveness, and stimulation.
\textit{Second}, participants completed a task-specific Human–Robot Interaction (HRI) questionnaire using 5-point Likert-scale ratings (defined from UEQ Dimensions). This questionnaire evaluated four dimensions which are summarized in Table~\ref{tab:hri_metrics}.

\begin{table}[htp]
\centering
\caption{Subjective HRI Questionnaire Dimensions}
\label{tab:hri_metrics}
\begin{tabular}{p{1.8cm} p{6.0cm}}
\toprule
\textbf{Dimension} & \textbf{Description} \\
\midrule
Perspicuity & Understanding of system behavior and awareness of ongoing task states. \\

Dependability & Perceived reliability and sense of control during execution. \\

Stimulation & Engagement and feeling accompanied during task progress. \\

Attractiveness & Overall enjoyment of the interaction. \\
\bottomrule
\end{tabular}
\end{table}

Participants also indicated their overall condition preference. Statistical comparisons were conducted between conditions to evaluate the impact of socially mediated state externalization on both objective performance and perceived interaction quality.

\textbf{Statistical Analysis.} All statistical analyses were conducted using paired methods appropriate for the within-subject design. For each objective performance metric (task initiation time, execution time, end-to-end time, grasp attempts, attention ratio, task success rates) and subjective rating (perspicuity, dependability, stimulation, attractiveness), we performed paired t-tests to compare conditions.
To account for potential order effects introduced by the counterbalanced design, we additionally fitted linear mixed-effects models for continuous outcome variables, with Condition (Hidden vs External) and Period (first vs second exposure) as fixed effects and participant as a random intercept. Effect sizes (Cohen's $d$) were computed to characterize the magnitude of observed differences. Finally, failure mode distributions were analyzed descriptively to verify that any subjective differences between conditions could not be attributed to systematic differences in underlying failure patterns. Statistical significance was evaluated at $\alpha = 0.05$.

\section{RESULTS}

This section reports objective performance metrics and subjective perception measures under both experimental conditions. 
All statistical comparisons were conducted using paired analyses due to the within-subject design ($N = 30$).

\subsection{Objective Performance Metrics}

\begin{figure}[t]
  \centering
  \includegraphics[width=0.99\linewidth]{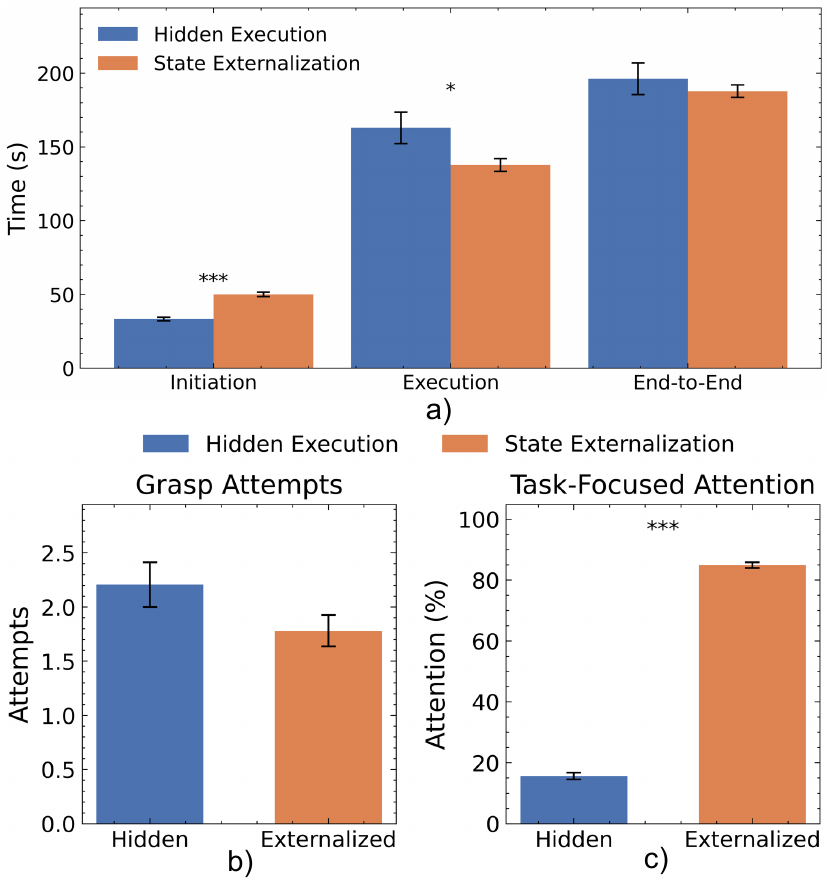}
  \caption{Objective performance and interaction metrics under Hidden Execution and Socially Mediated State Externalization. 
(a) Time-based metrics, including task initiation time, execution time, and total end-to-end duration. 
(b) Average number of grasp attempts per task session. 
(c) Task-focused attention ratio during execution. \textit{Error bars represent $\pm$1 SEM. 
Asterisks denote statistical significance ($^*p<0.05$, $^{***}p<0.001$).}.}
  \label{fig:objfig}
\end{figure}

\begin{figure}[!htp]
  \centering
  \includegraphics[width=0.99\linewidth]{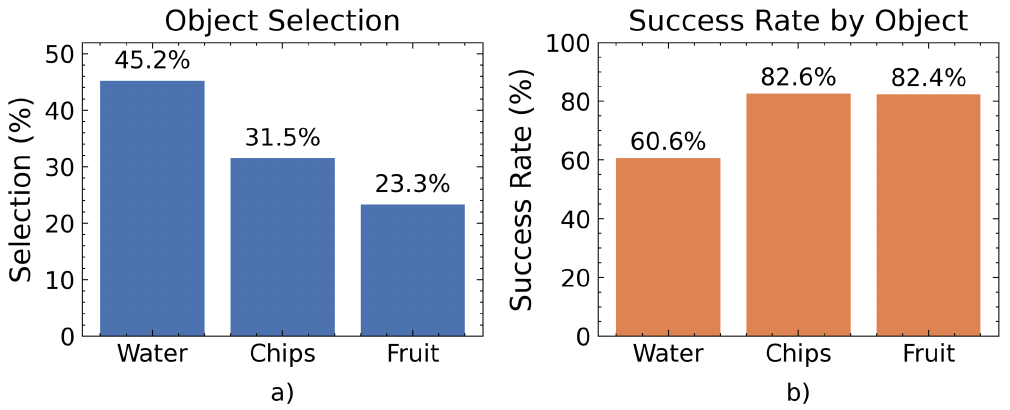}
  \caption{
Object-level task distribution and manipulation success rates. 
(a) Percentage of object requests across categories (Water, Chips, Fruit). 
(b) Object-wise task success rates aggregated across conditions. 
Although success rates varied across object types, object requests were moderately distributed and no condition-specific interaction was observed, indicating that object composition did not confound the transparency effect.}
  \label{fig:stat}
\end{figure}

We first examine whether socially mediated state externalization affected system-level execution performance (Fig.~\ref{fig:objfig} \&  Table~\ref{tab:objmetrics}). To ensure that performance differences were not driven by task selection bias or object-specific difficulty, we examined the distribution of object requests and object-wise success rates (Fig.~\ref{fig:stat}). Object selections were relatively balanced across categories. Success rates were comparable across object types and did not show systematic differences between conditions, suggesting that the observed perceptual improvements are unlikely to be attributable to object-specific performance variation. Overall, objective metrics indicate that socially mediated state externalization significantly increased user attention while preserving overall task efficiency and physical execution performance.

\begin{table}[t]
\centering
\setlength{\tabcolsep}{3pt}
\caption{Objective performance under Autonomous Hidden Execution (A) and Socially Mediated State Externalization (B).}
\label{tab:objmetrics}

\begin{tabular}{lcccc}
\toprule
Metric & A Mean (SD) & B Mean (SD) & t(29) & p \\
\midrule
Init. (s) & 33.47 (7.84) & 49.93 (9.12) & 9.21 & $<0.001^{***}$ \\
Exec. (s) & 162.63 (64.54) & 137.33 (35.87) & -2.56 & $0.016^{*}$ \\
Total (s) & 196.10 (67.78) & 187.26 (38.40) & -1.12 & $0.271$ \\
Grasp Att. & 2.20 (1.21) & 1.78 (0.93) & -1.54 & $0.134$ \\
Attention (\%) & 15.8 (8.3) & 84.6 (6.5) & 21.47 & $<0.001^{***}$ \\
Suc. Rate (\%) & 86.7 & 90.0 & -- & $0.754$ \\
\bottomrule
\end{tabular}
\end{table}

\subsection{Subjective Perception Measures}


Subjective ratings collected through the UEQ-inspired HRI questionnaire are summarized in Table~\ref{tab:subjective}. Across all four dimensions—perspicuity, dependability, stimulation, and attractiveness—participants reported significantly higher scores under the Socially Mediated State Externalization condition. Participants’ overall preference also favored the socially mediated state externalization condition: 83\% selected the externalized system, 10\% preferred the hidden execution baseline, and 7\% reported no preference.

\begin{table}[htp]
\centering
\small
\setlength{\tabcolsep}{3pt}
\caption{Subjective ratings for Autonomous Hidden Execution (A) vs Socially Mediated State Externalization (B).}
\label{tab:subjective}
\begin{tabular}{lcccc}
\toprule
Metric & A Mean (SD) & B Mean (SD) & t(29) & p \\
\midrule
Perspicuity & 3.12 (0.74) & 4.31 (0.62) & 6.84 & $<0.001^{***}$ \\

Dependability & 3.36 (0.69) & 4.18 (0.58) & 4.97 & $<0.001^{***}$ \\

Stimulation & 3.08 (0.77) & 4.45 (0.51) & 8.12 & $<0.001^{***}$ \\

Attractiveness & 3.24 (0.72) & 4.36 (0.55) & 6.21 & $<0.001^{***}$ \\

\bottomrule
\end{tabular}
\end{table}






Collectively, these results demonstrate that socially mediated state externalization significantly enhanced perceived transparency, engagement, and overall user experience without compromising objective task performance.

\subsection{Order and Execution-Aware Analysis}

To account for potential order effects in the counterbalanced within-subject design, we fitted linear mixed-effects models with participant as a random intercept and fixed effects for Condition (Hidden vs.\ External) and Period (first vs.\ second exposure). Results are summarized in Table~\ref{tab:lmm}.

\begin{table}[t]
\centering
\footnotesize
\setlength{\tabcolsep}{3pt}
\renewcommand{\arraystretch}{0.95}
\caption{Linear mixed-effects analysis of order effects. Fixed effects include Condition (Hidden vs.\ External) and Period (first vs.\ second exposure); participant is a random intercept.}
\label{tab:lmm}
\begin{tabular}{lcccc}
\toprule
Outcome & Effect & $\beta$ & Test stat. & $p$ \\
\midrule
\multirow{2}{*}{Total task time} 
  & Condition & 7.98 & $z=0.70$ & 0.484 \\
  & Period    & -2.85 & -- & 0.803 \\
\midrule
\multirow{2}{*}{Grasp attempts} 
  & Condition & 0.44 & $z=1.89$ & 0.059 \\
  & Period    & -0.26 & -- & 0.274 \\
\bottomrule
\end{tabular}
\end{table}

\begin{figure}[htp]
  \centering
  \includegraphics[width=0.99\linewidth]{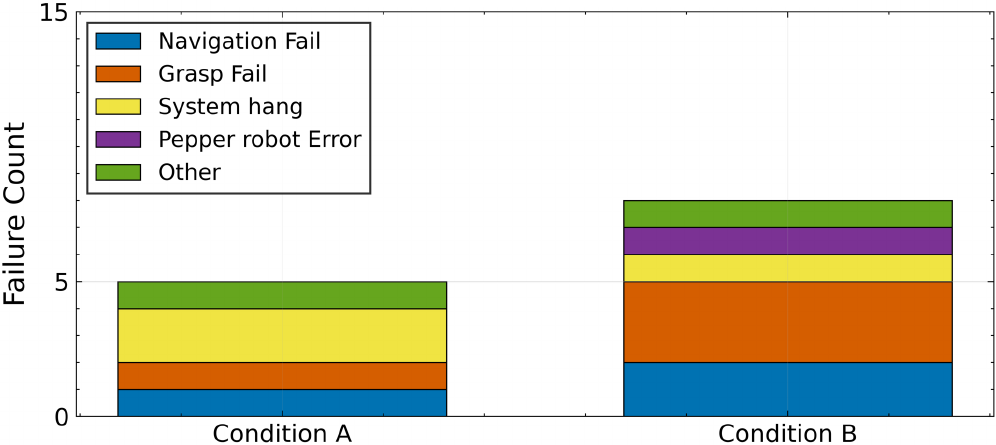}
  \caption{Failure mode breakdown for failed trials under both experimental conditions. 
Failure events were categorized into navigation errors, grasp failures, system hangs, Pepper-related errors, and other/unknown causes. 
Although the total number of failed trials differed slightly, the relative composition of failure types remained comparable across conditions, indicating no systematic shift in underlying failure mechanisms.}
  \label{fig:failed}
\end{figure}

\textbf{Failure modes.} We additionally examined the composition of failure modes across conditions (Fig.~\ref{fig:failed}) and did not observe a condition-specific concentration in any single failure category.

Overall, these results indicate that the observed subjective improvements under socially mediated state externalization cannot be attributed to order effects or systematic differences in execution efficiency. These findings strengthen the interpretation that improvements in perceived transparency and engagement arise from the state externalization mechanism itself rather than from confounding performance differences.

\section{DISCUSSION}


\subsection{Impact of State Externalization}

Our results show that real-time socially mediated state externalization substantially improves user perception in distributed home robot settings where execution occurs outside the user’s field of view. Compared to Autonomous Hidden Execution, socially mediated state externalization significantly increased subjective ratings of perspicuity, dependability, stimulation, and attractiveness (all $p<.001$), and a large majority of participants (83\%) preferred the externalized condition.

Beyond self-report, we observed a strong behavioral effect: participants’ task-focused attention increased from 15.8\% to 84.6\% ($p < .001$). This suggests that state awareness mechanisms do not merely change what users \emph{think} about the system, but also reshape how users \emph{stay engaged} with the task loop when execution is physically separated. In the hidden condition, the absence of visible progress cues likely encouraged disengagement (e.g., shifting attention to unrelated activities), whereas the mediator’s verbal and visual updates anchored attention to task-relevant feedback.

Importantly, these perceptual and behavioral benefits cannot be explained by confounding performance differences or order effects. Mixed-effects analyses accounting for period and repeated measures showed that overall completion time did not differ reliably between conditions, supporting the interpretation that the transparency mechanism itself drove the observed improvements.

\subsection{Transparency Without Compromising Autonomy}

A common concern in HRI is that increasing system transparency may 
introduce interaction overhead that undermines autonomous operation. 
Our results suggest that this need not be the case: socially mediated 
state externalization achieved significant gains in user understanding 
and engagement while preserving overall task efficiency.

Concretely, state externalization increased task initiation time 
(33.47 s vs. 49.93 s, $p < .001$), reflecting the additional interaction 
turns required to establish shared understanding through confirmations 
and explicit progress updates. However, this upfront investment in 
transparency did not translate into longer overall task duration: 
end-to-end time remained comparable across conditions ($p = .271$). 
Notably, execution time was actually reduced under state 
externalization (162.63 s vs. 137.33 s, $p = .016$).

One plausible explanation for this pattern emerges from examining 
grasp attempt data. Although the difference was not statistically 
significant ($p = .134$), we observed more grasp attempts on average 
in the Hidden condition (M = 2.20) than in the Externalized condition 
(M = 1.78). This suggests that the execution robot may have 
experienced more frequent manipulation difficulties in Condition A, 
potentially due to uncontrolled environmental factors or the inherent 
unpredictability of grasping tasks. These additional recovery attempts 
would increase execution time in the Hidden condition, helping to 
explain why the longer initiation time in the Externalized condition 
did not result in longer end-to-end duration. In effect, the two 
conditions incurred different types of time costs—interaction overhead 
in one case, recovery overhead in the other—that balanced out in the 
final measure.

Critically, these objective performance differences do not undermine 
the primary finding: socially mediated state externalization 
dramatically improved perceived transparency, dependability, and 
engagement (all $p < .001$) without sacrificing overall task efficiency. 
Participants in the Externalized condition remained actively engaged 
throughout (task-focused attention: 84.6\% vs. 15.8\%, $p < .001$), 
suggesting that transparency mechanisms can keep users appropriately 
informed without introducing prohibitive time costs.

These findings demonstrate that transparency and autonomy need not be 
competing objectives. When implemented through appropriate social 
mediation, state externalization can enhance the human-robot 
partnership while maintaining the efficiency gains that autonomous 
operation provides.

\subsection{Implications for Distributed Home Robotics}

Our findings suggest several design implications for future distributed domestic robot ecosystems. \textbf{(1) Externalize task states when execution is out-of-sight.}
When users cannot directly observe robot motion, the system loses implicit behavioral cues that normally support predictability. In such contexts, task-level state externalization (e.g., NAVIGATING, GRASPING, RECOVERING) becomes a functional substitute for visual observability and substantially improves perceived understanding. Our results confirm that even simple verbal and visual updates can transform a system that feels opaque into one that feels predictable and dependable. \textbf{(2) Communicate task-level semantics, not low-level telemetry.}
Exposing raw sensor data or low-level controller signals would likely overwhelm users and increase cognitive workload. Instead, a compact set of interpretable task states provides actionable awareness without requiring users to become robot experts. This suggests that designing appropriate abstraction layers is a system-level requirement for scalable transparency -- one that our socially mediated approach successfully implemented. \textbf{(3) Leverage physically-present social mediation to bridge the awareness gap.}
A co-located social mediator does more than translate internal execution processes into human-readable updates through speech and lightweight visualization. Unlike a phone or tablet interface, the mediator's physical embodiment and presence and social expressiveness create a sense of co-presence and companionship during what would otherwise be a solitary waiting period. In our study, Pepper served not merely as a display for task state information, but as an engaging social actor that anchored user attention, acknowledged progress, and requested confirmation when needed. This physical-social presence transformed the experience from passively awaiting an autonomous system's return to actively collaborating with a visible, communicative partner. The dramatic increase in task-focused attention (from 15.8\% to 84.6\%) suggests that physically-present social mediation engages users in ways that purely graphical interfaces cannot. \textbf{(4) Treat transparency as an architectural variable, not a purely functional add-on.}
In distributed systems, decisions about when to externalize failures, when to request user confirmation, and when to silently recover shape trust and perceived reliability. These are not merely surface-level presentation choices, but system-level policies that should be designed alongside task execution logic. Our experimental manipulation varied exactly such policies, and the resulting perceptual gains highlight the importance of embedding transparency into the core system architecture.

As multi-room, multi-agent home robot deployments become more common, mitigating the state awareness gap may be essential for maintaining user trust and sustained engagement, rather than an optional enhancement. Overall, the results suggest that socially mediated state externalization provides a high-leverage design intervention: it yields large gains in perceived transparency and engagement while maintaining comparable end-to-end performance.

\section{LIMITATIONS AND FUTURE WORK}

This study has several limitations that should be considered when interpreting the results.

\textbf{First}, participants were healthy adults (18+), recruited from local universities and the surrounding community (including working professionals), who evaluated the system in a short-term laboratory setting. Although this controlled design enables rigorous within-subject comparison, it does not fully capture long-term domestic use, assistive dependency, or real household routines. In real homes, repeated interaction may influence expectations of transparency, tolerance for recovery delays, and reliance on mediator feedback. Longitudinal in-situ deployments are needed to examine how state externalization affects trust calibration, sustained engagement, and reliance patterns over time. \textbf{Second}, the evaluated task domain was limited to single-object retrieval and delivery with a relatively constrained set of execution states and failure modes. More complex multi-step activities—such as sequential manipulation, multi-room coordination, or collaborative tasks requiring ongoing human input—may generate richer uncertainty and more frequent state transitions. The scalability of task-level semantic state abstraction and externalization strategies should therefore be validated in more diverse and cognitively demanding household scenarios. \textbf{Third}, the study compared two boundary conditions: fully hidden execution and full real-time state externalization. Intermediate transparency designs (e.g., failure-only notifications, delayed updates, or passive visual indicators without verbal mediation) were not examined. It is possible that graded transparency levels could achieve similar perceptual benefits while reducing interaction overhead. Systematic exploration of partial or adaptive transparency mechanisms would clarify how much state exposure is necessary to maintain predictability and engagement. \textbf{Fourth}, although end-to-end task duration remained statistically comparable across conditions, state externalization increased task initiation time due to additional interaction turns for confirmation and state framing. While this overhead did not degrade overall completion efficiency in our controlled setting, its impact on cognitive workload and perceived efficiency in time-critical or high-frequency task contexts remains unclear. Future work should incorporate explicit workload measures (e.g., NASA-TLX) to quantify these effects. \textbf{Finally}, task-focused attention was derived from structured video-based behavioral coding. While this metric provided objective evidence of engagement differences, higher-resolution sensing methods such as eye-tracking or physiological measures could further strengthen behavioral interpretation and reduce potential coding bias.

\textbf{Future research} should investigate adaptive transparency mechanisms that dynamically adjust the level and timing of state externalization based on contextual factors such as task criticality, failure likelihood, detected user attention state, or spatial separation. Rather than treating transparency as a static design choice, adaptive mediation strategies may enable distributed robotic systems to balance autonomy and user awareness more effectively across diverse domestic environments.

\section{CONCLUSION}

Distributed home robot systems increasingly decouple user interaction from physical task execution, creating a state awareness gap when robot activity occurs outside the user’s field of view. This work investigated whether real-time, socially mediated state externalization can bridge this gap without degrading task performance.

Through a controlled within-subject study ($N=30$), we showed that externalizing task-level execution states significantly enhances perceived transparency, dependability, stimulation, and overall attractiveness, while substantially increasing task-focused attention. Importantly, these perceptual and behavioral improvements were achieved without increasing overall end-to-end task duration, and could not be attributed to order effects or execution variability. The findings suggest that transparency in distributed robotics should be treated as an architectural design variable rather than a purely functional output feature. When execution is spatially separated from users, externalizing semantically meaningful task states can meaningfully improve user understanding and engagement, even when underlying autonomy remains unchanged. As domestic robot ecosystems become increasingly multi-room and multi-agent, designing mechanisms that mitigate the state awareness gap will be essential for maintaining user confidence and sustained interaction. Socially mediated state externalization represents one effective step toward this goal.

\addtolength{\textheight}{-12cm}   








\bibliographystyle{IEEEtran}
\bibliography{references}





\end{document}